\documentclass{article}
\pdfoutput=1
\usepackage{spconf,amsmath,graphicx} 

\usepackage{enumitem}
\setlist{nosep, leftmargin=14pt}

\usepackage{mwe} 
\usepackage{booktabs} 
\usepackage{multirow}
\usepackage{tabularx}
\usepackage{adjustbox}
\usepackage{array}
\usepackage{svg}
\usepackage{gensymb}
\usepackage{anyfontsize}
\usepackage{amssymb}
\usepackage{pifont}
\usepackage{algorithm,algorithmicx}
\usepackage[noend]{algpseudocode}

\usepackage{makecell}

\title{ConvFormer: Combining CNN and Transformer for Medical Image Segmentation}
\name{Pengfei Gu$^{\dagger}$ 
     \quad Yejia Zhang$^{\dagger}$ 
     \quad Chaoli Wang$^{\dagger}$
     \quad Danny Z. Chen$^{\dagger}$ 
     }
\address{ $^{\dagger}$University of Notre Dame, Department of Computer Science and Engineering, Notre Dame, IN, USA}
\begin{document}
%

\maketitle
\begin{abstract}
Convolutional neural network (CNN) based methods have achieved great successes in medical image segmentation, but their capability to learn global representations is still limited due to using small effective receptive fields of convolution operations.
Transformer based methods are capable of modelling long-range dependencies of information for capturing global representations,
yet their ability to model local context is lacking.
Integrating CNN and Transformer to learn both local and global representations while exploring multi-scale features is instrumental in further improving medical image segmentation.
In this paper, we propose a hierarchical CNN and Transformer hybrid architecture, called ConvFormer, for medical image segmentation. ConvFormer is based on several simple yet effective designs.
(1) A feed forward module of Deformable Transformer (DeTrans) is re-designed to introduce local information, called Enhanced DeTrans. 
(2) A residual-shaped hybrid stem based on a combination of convolutions and Enhanced DeTrans is developed to capture both local and global representations to enhance representation ability. 
(3) Our encoder utilizes the residual-shaped hybrid stem in a hierarchical manner to generate feature maps in different scales, and an additional Enhanced DeTrans encoder with residual connections is built to exploit multi-scale features with feature maps of different scales as input. 
Experiments on several datasets show that our ConvFormer, trained from scratch, outperforms various CNN- or Transformer-based architectures, achieving state-of-the-art performance.
\end{abstract}
%
%
\vspace{-1mm}
\section{Introduction}\label{sec:intro}
\vspace{-1mm}
Image segmentation is a central problem in medical image analysis. Convolutional neural networks (CNNs), especially fully convolutional networks (FCNs), have become predominant approaches for this problem (e.g., U-Net~\cite{ronneberger2015u}, UNet 3+~\cite{huang2020unet}, etc).
Despite their successes, CNNs still have yet to address the limitation in learning long-range dependencies (global information) to see a ``big picture'' due to limited effective receptive fields (ERFs) of convolution (Conv) operations.  
Attempts were made to enlarge ERFs, such as utilizing 
atrous Convs with different dilated rates (e.g.,~\cite{chen2017deeplab}), applying pyramid pooling (e.g.,~\cite{zhao2017pyramid}), and designing large kernels (e.g.,~\cite{peng2017large}).
%
Although these methods enlarged ERFs to some extent, they still suffered from limited ERFs, yielding sub-optimal image segmentation accuracy. 

Recently, Transformers (e.g., vision Transformer~\cite{dosovitskiy2020image}) became a de-facto choice for modelling long-range dependencies in computer vision, inspired by their success with self-attention mechanism in natural language processing. 
Compared to CNN methods, Transformer models have larger receptive fields and excel at learning global information. But, they also have drawbacks, e.g., high computation cost, slow convergence, and short of CNN's inductive biases.
Two types of methods attempted to reduce their computation cost. (1) Limiting self-attention to local windows (e.g.,~\cite{liu2021swin,zhu2020deformable}).
%
(2) Downsampling the key and value feature maps (e.g.,~\cite{wang2021pyramid}). 
Though effective in capturing global information, these Transformer-based methods still yielded unsatisfactory performance due to deficiency in learning local information.

In medical image segmentation, efforts were made to combine Transformer and CNN to learn both local and global representations. In~\cite{chen2021transunet}, Transformer was utilized as a supplement, appended to the last Conv block of the CNN encoder to learn global information.
MedT~\cite{valanarasu2021medical} exploited local and global information by employing two branches, where a gated axial Transformer was applied to explore global information and CNN was leveraged to learn local information.
CoTr~\cite{xie2021cotr} employed Deformable Transformer (DeTrans) as an additional encoder for exploring multi-scale information from multi-scale feature maps for 3D medical image segmentation.
MISSFormer~\cite{huang2021missformer} introduced a Conv layer to Transformer to enhance its capability to learn local information. 
UNETR~\cite{hatamizadeh2022unetr} proposed a U-shaped encoder-decoder network, where Transformer was used as the encoder to capture global multi-scale information and a CNN decoder was used to compute final segmentation output.  
Although achieving promising performances, these methods still incurred several drawbacks.
(1) Limited capability to learn both local and global representations due to ineffective integration of CNN and Transformer. 
For example, CoTr~\cite{xie2021cotr} did not apply Conv to DeTrans for effectively learning local information. TransUNet~\cite{chen2021transunet} simply appended Transformer to a Conv block in the encoder. UNETR~\cite{hatamizadeh2022unetr} did not leverage Conv in the encoder.
(2) Deficiency in capturing multi-scale information. For instance, MedT~\cite{valanarasu2021medical} did not explore multi-scale information in its global Transformer branch, 
and such drawbacks in learning both local and global representations and capturing multi-scale information led to sub-optimal segmentation. 

In this paper, we propose a new hierarchical CNN and Transformer hybrid architecture, called ConvFormer, to capture both local and global representations while exploiting multi-scale features for medical image segmentation. Specifically, ConvFormer is based on several key designs.
(1) We re-design the feed forward module of DeTrans to introduce local information, called Enhanced DeTrans. 
(2) We develop a residual-shaped hybrid stem based on a combination of Convs and Enhanced DeTrans to capture both local and global representations to enhance representation ability. 
(3) The residual-shaped hybrid stem is utilized in the encoder in a hierarchical manner to generate feature maps in different scales, and an additional Enhanced DeTrans encoder with residual connections is built to explore multi-scale features with feature maps of different scales as input. 
(4) An enhanced positional encoding (EPE) introduces a Conv layer to the fixed sinusoidal encoding method~\cite{vaswani2017attention} to improve adaptability and flexibility.

Extensive experiments on two public datasets (2017 ISIC Skin Lesion segmentation (2D)~\cite{codella2018skin} and MM-WHS CT (3D)~\cite{zhuang2016multi}) and one in-house dataset (lymph node segmentation (2D)) show that our ConvFormer, trained from scratch, outperforms various known CNN-based or Transformer-based methods, achieving state-of-the-art performance.

\begin{figure}[t]
    \centering
\includegraphics[width=1.0\linewidth]{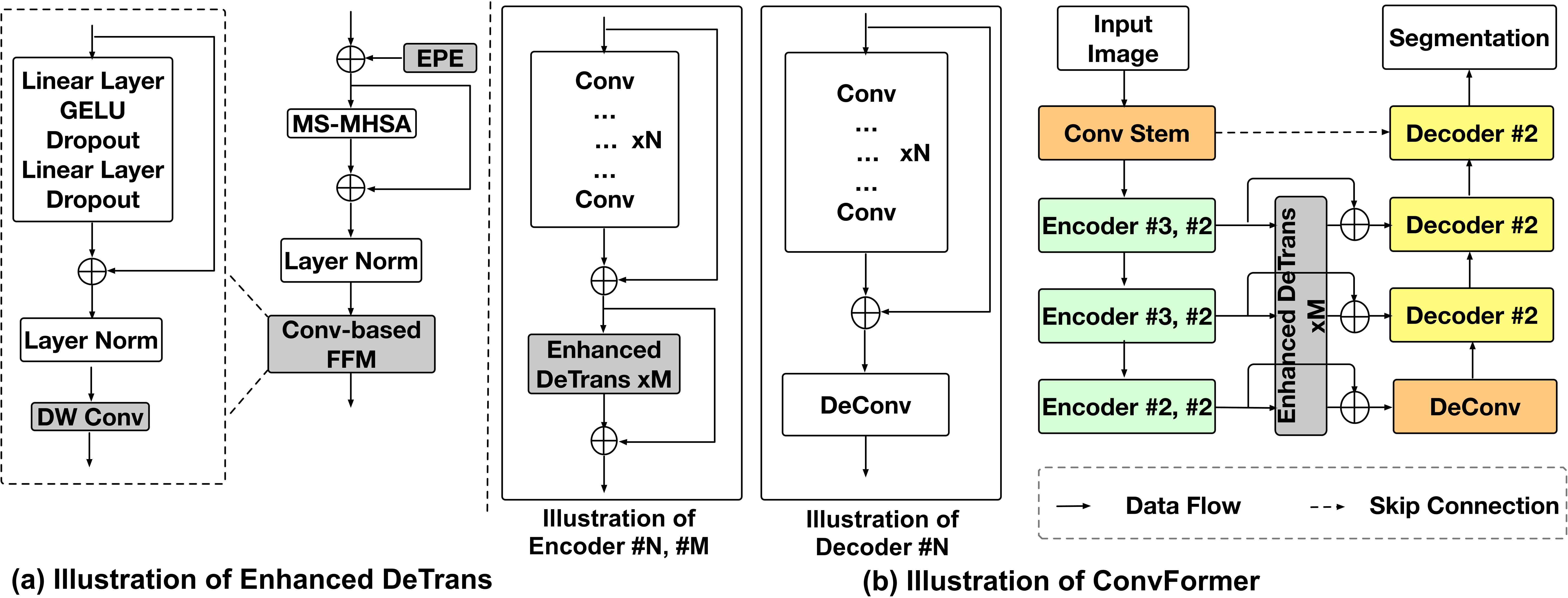}
\vspace{-8mm}
    \caption{(a) An overview of our Enhanced DeTrans with the proposed enhanced positional encoding (EPE) and Conv-based feed forward module (FFM) (shown in gray color). (b) Our ConvFormer: The two big boxes on the left are for the encoder and decoder, respectively (best viewed in color).}
    \label{fig:archi}
    \vspace{-3mm}
\end{figure}
\vspace{-2mm}
\section{Method}\label{sec:method}
\vspace{-2mm}
Fig.~\ref{fig:archi} shows the architecture of our ConvFormer that contains three main components: (1) Enhanced DeTrans that captures global representations with local information, using a Conv-based feed forward module (FFM); 
(2) the residual-shaped hybrid stem that extracts local and global representations in different scales that are fed to an additional Enhanced DeTrans encoder to explore multi-scale features; 
(3) enhanced positional encoding (EPE) that improves the adaptability and flexibility of the fixed sinusoidal encoding method~\cite{vaswani2017attention}.

\vspace{-3mm}
\subsection{Enhanced DeTrans}\label{subsec:ffm}
\vspace{-1mm}
FFM is an essential component of Transformer to enhance the representation ability. The FFM proposed in DeTrans~\cite{zhu2020deformable,xie2021cotr} consists of a multilayer perceptron (MLP) composing of two linear layers separated by GELU, as follows:
\begin{equation}\label{eq-FFM}
    \text{FFM}(x) = \text{LN}\left(\text{GELU}(xW_1 + b_1)W_2 + b_2 + x\right),
\end{equation}
where $x$ is an input feature map, $W_1$ and $W_2$ are weights of the two linear layers respectively, $b_1$ and $b_2$ are bias terms, and $\text{LN}$ denotes layer norm. For simplicity, Dropouts are omitted. 
Though effective, this design does not have good capability to learn local information, which is critical in dense predictions.

To overcome this limitation and enable FFM to introduce local information to the global representations captured by multi-scale multi-head self attention (MS-MHSA), we resort to a Conv layer.
Specifically, we insert depth-wise convolution (DW Conv) to the end of FFM. The resulted FFM is called Conv-based FFM.
Note that we leverage DW Conv instead of Conv for reducing computation cost.
As a result, the proposed Conv-based FFM inherits the merit of both CNN and DeTrans in learning local and global representations.

To adapt DW Conv to FFM, we first reshape the input 1D sequences captured by FFM to 2D/3D feature maps, apply DW Conv to the reshaped feature maps to learn local information, and reshape the result back to 1D sequences as output:
\begin{align} 
\label{eq-Conv-based-FFM}
\text{Conv-based FFM}(x) &=  \text{reshape}\left(\text{DW Conv}\left(\text{reshape}(\text{FFM}(x))\right)\right),
\end{align}
where $x$ is an input feature map. Note that our Conv-based FFM is capable of dealing with multi-scale feature map input, which is important for accurate segmentation. DW Conv is shared when processing multi-scale feature maps.

We replace the FFM of DeTrans by our Conv-based FFM, and the resulted DeTrans is called Enhanced DeTrans. Enhanced DeTrans is capable of capturing both local and global information by combining Conv and Transformer.

\vspace{-3mm}
\subsection{Residual-shaped Hybrid Stem}\label{subsec:hybrid-stem}
\vspace{-1mm}
Stacking Transformer to the last Conv block as a global feature extractor is a common way to combine CNN/Transformer in encoder to learn 
local and global representations. 
For example, in TransUNet~\cite{chen2021transunet} and CoTr~\cite{xie2021cotr}, Transformer was followed by a Conv block to capture global information for local features obtained by Conv blocks.
Although such methods might enhance the models' representation ability on learning local and global information to some extent, they did not exploit both local and global features effectively. 
More importantly, they did not generate and explore multi-scale local and global features, which are highly important to handle datasets with large variations in object size, shape, and texture, which are common characteristics of medical image datasets.  

To effectively capture both local and global representations, we propose the residual-shaped hybrid stem, which consists of two key designs (see the leftmost box in Fig.~\ref{fig:archi}(b)).
(1) Residual connections are introduced to the stacked Convs for feature diversity and delivery. (2) Local representations captured by the stacked Convs are combined with global representations learned by Enhanced DeTrans via an add operation. 
These two designs offer two compelling advantages: (i) The introduced residual connections are critical for avoiding feature collapse (i.e., feature diversity is continuously reduced as the layer depth increases);
(ii) the captured local and global representations can be better fused via add operation. 
The improvement (e.g., in Table~\ref{tab:ablation}, F1 improved by $1.0\%$, $p<0.05$, t-test) shows the importance of the introduced residual connections.

%
We use the residual-shaped hybrid stem in the encoder in a hierarchical manner, so that it can generate global and local representations in different scales. 
Given an input image, we can obtain hierarchical global and local representations with different resolutions, which are then fed to an additional Enhanced DeTrans encoder to exploit multi-scale representations. 
Different from CoTr~\cite{xie2021cotr} which utilized DeTrans to exploit multi-scale features, we add residual connections to our Enhanced DeTrans to reuse the local representations captured by the stacked Convs and consolidate the captured local and global representations.

The architecture of our ConvFormer is shown in Fig.~\ref{fig:archi}. Following the architecture of CoTr~\cite{xie2021cotr}, ConvFormer has four encoder and decoder stages, with a Conv stem and three residual-shaped hybrid stems in encoder, a deconvolution (DeConv) stem and three decoder stems in decoder, and an additional Enhanced DeTrans encoder. 

\vspace{-3mm}
\subsection{Enhanced Positional Encoding (EPE)}\label{subsec:hybrid-EPE}
\vspace{-1mm}
In Transformer, we first flatten input feature maps into 1D sequences. But, this flattening process may cause loss of spatial information that is critical for segmentation. 
Known methods~\cite{xie2021cotr,vaswani2017attention} employed fixed sinusoidal encoding to supplement flattened sequences with position information. 
In particular, sine and cosine functions of different frequencies were used to compute the positional coordinates of each dimension:
\begin{align}\label{eq-fixed-PE}
\text{PE}_{(pos, 2i)} &= \sin\left(pos/10000^{2i/C}\right), \\ 
\text{PE}_{(pos, 2i+1)} &= \cos\left(pos/10000^{2i/C}\right), 
\end{align}
where $pos$ is for position, $i$ is the dimension, and $C$ is a constant. 
However, such a fixed sinusoidal encoding method lacks adaptability and flexibility, as the code (e.g., frequencies) is predefined. 

To improve adaptability and flexibility, we introduce DW Conv to the fixed sinusoidal encoding~\cite{vaswani2017attention}, since a Conv kernel naturally encodes pixel locality and semantic continuity
with adaptability and flexibility. 
The resulted positional encoding method is called enhanced positional encoding (EPE). 
Specifically, it uses two branches: one branch applies the fixed sinusoidal encoding~\cite{vaswani2017attention} to learn position information, and the other uses DW Conv to capture position information. The two branches are combined by an add operation:
\begin{equation}\label{eq-EPE}
    x' = \text{fixed sinusoidal encoding}(x) + \text{ReLU}(\text{BN}(\text{DW Conv}(x))),
\end{equation}
where $x$ is an input feature map, $x'$ denotes the output feature map embedded with position information, and $\text{BN}$ is batch normalization.
The improvement (in Table~\ref{tab:ablation}, F1 improved by $0.4\%$, $p<0.05$, t-test) demonstrates the importance of EPE.

\vspace{-2mm}
\section{Experiments and Results}\label{sec:exp}
\vspace{-2mm}
{\bf Datasets.} (1) {\bf The lymph node segmentation ultrasound dataset:} 
This in-house dataset is for segmenting lymph nodes in 2D ultrasound images.
It contains 137 training and 100 test images.
(2) {\bf The 2017 MM-WHS CT dataset:} 
This public dataset~\cite{zhuang2016multi} is for segmenting seven cardiac structures, the left/right ventricle blood cavity (LV/RV), left/right atrium blood cavity (LA/RA), myocardium of the left ventricle (LV-myo), ascending aorta (AO), and pulmonary artery (PA). It contains 20 unpaired 3D CT images, which are randomly split into 16 images and 4 images for training and testing, following~\cite{zheng2019hfa}.
(3) {\bf The 2017 ISIC skin lesion segmentation dataset:} 
This public dataset~\cite{codella2018skin} is for segmenting lesion boundaries in 2D dermoscopic images. It contains 2000 training, 150 validation, and 600 test images. 

\noindent 
{\bf Implementation Details.}
Our ConvFormer is implemented with PyTorch, and is trained on an NVIDIA Tesla P100 Graphics Card with 16GB GPU memory using the AdamW~\cite{loshchilov2017decoupled} 
optimizer with weight decay $0.005$. 
We apply the ``poly" learning rate policy with an initial learning rate of $2e-4$. The maximum number of iterations is $100$k for the lymph node and 2017 ISIC skin lesion datasets, and $240$k for the 2017 MM-WHS CT dataset (using about $34$, $48$, and $96$ training hours, respectively).
To avoid overfitting, standard data augmentation (e.g., random flip, crop, etc) is applied.

\begin{table}[t]
    \centering
    \caption{Quantitative results of various models on the lymph node dataset. The reported values are average$\pm$STD for 3 runs with different random seeds. 
    The best results are in {\bf bold}.
    }
    \label{tab:lymph}
    \scriptsize
    \resizebox{\columnwidth}{!}
    {
    \begin{tabular}{|ccccc|}
        \hline
        Method &IoU  &Precision  &Recall  &F1 Score\\
        \hline
        U-Net~\cite{ronneberger2015u}& 0.661  & 0.834  & 0.761  & 0.796\\\hline
        Zhang et al.~\cite{zhang2019decompose}& 0.810 & 0.901 & 0.889  &0.895\\
        \hline
        $k$CBAC-Net~\cite{gu2021kcbac} &0.829   &0.909  &0.904 &0.906   \\\hline
        ConvFormer (Ours) & \textbf{0.845}$\pm$0.002   &\textbf{0.925}$\pm$0.002  &\textbf{0.907}$\pm$0.002  &\textbf{0.916}$\pm$0.002    \\\hline
    \end{tabular}
}
\vspace{-4mm}
\end{table}

\begin{figure}[h]
    \centering
    \includegraphics[width=1.0\linewidth]{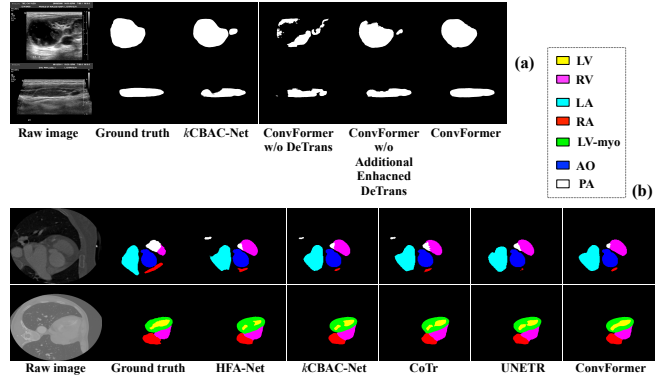}
    \vspace{-8mm}
    \caption{Some visual qualitative results on the lymph node dataset (a) and the 2017 MM-WHS CT dataset (b), demonstrating the capability of our ConvFormer.
    }
    \label{fig:vis-result}
    \vspace{-4mm}
\end{figure}

\noindent 
{\bf Experimental Results.}
Table~\ref{tab:lymph} reports quantitative results on the lymph node dataset, showing that our ConvFormer outperforms the known methods by a clear margin in all the evaluation measures. 
In particular, ConvFormer outperforms the state-of-the-art (SOTA) method, $k$CBAC-Net~\cite{gu2021kcbac}, by 1.6\% and 1.0\% in IoU and F1, respectively, achieving new SOTA performances.
Table~\ref{tab:whs} shows quantitative results on the 2017 MM-WHS CT 3D dataset. Our ConvFormer outperforms the best-known CNN-based (HFA-Net~\cite{zheng2019hfa}, $k$CBAC-Net~\cite{gu2021kcbac}) and Transformer-based (CoTr~\cite{xie2021cotr}, UNETR~\cite{hatamizadeh2022unetr}) methods, achieving new SOTA performances.
Specifically, ConvFormer outperforms HFA-Net~\cite{zheng2019hfa}, $k$CBAC-Net~\cite{gu2021kcbac}, CoTr~\cite{xie2021cotr}, and UNETR~\cite{hatamizadeh2022unetr} by 1.6\%, 0.6\%, 1.4\%, and 1.0\% in Dice score on average, respectively.
Table~\ref{tab:skin} gives quantitative results on the 2017 ISIC skin lesion dataset. Our ConvFormer attains accuracy gain of 3.2\% in Jaccard Index over the 2017 ISIC Challenge winner~\cite{yuan2017improving} and slightly outperforms the SOTA method, $k$CBAC-Net~\cite{gu2021kcbac}, achieving new SOTA performances. 
Note that, (1) our ConvFormer shows excellent generalization on both 2D and 3D datasets, on ultrasound, dermoscopic, and CT images, outperforming previous SOTA methods; 
(2) ConvFormer achieves competitive performances without pre-training, different from most of the known Transformer-based models. 
All these suggest that our ConvFormer is a promising and robust method suitable for medical image segmentation tasks on datasets of different imaging characteristics and modalities, and
it is capable of learning effective local, global, and multi-scale representations.  Fig.~\ref{fig:vis-result} shows some qualitative results.

\begin{table}[t]
    \centering
    \caption{Quantitative results of different models on the 2017 MM-WHS CT dataset. ``---'' means that the results were not reported by the original papers, ``Para.'' means the number of parameters of the model, and ``HDF'' represents Hausdorff.}
    \label{tab:whs}
    \scriptsize
    \resizebox{\columnwidth}{!}
    {
        \begin{tabular}{|ccccccccccc|}
            \hline
            {Method}&{Para.}&{Metric}&{LV}&{RV}&{LA}&{RA}&{\makecell{LV\\-myo}}&{AO}&{PA}&{Mean}\\
            \hline
            \multirow{1}{*}{Payer et al.~\cite{payer2017multi}}    
                     &---   & Dice   & 0.918 &0.909 &0.929 &0.888 &0.881 &0.933 &0.840 &0.900          \\
                        \hline
            \multirow{1}{*}{Dou et al.~\cite{dou2018unsupervised}}    
                       &---   & Dice   & 0.888 &--- &0.891 &--- &0.733 &0.813 &--- &---          \\
                        \hline
            \multirow{1}{*}{Chen et al.~\cite{chen2020unsupervised}}    
                     &---    & Dice   & 0.919 &--- &0.911 &--- &0.877 &0.927 &--- &0.909          \\
                        \hline
            \multirow{4}{*}{HFA-Net~\cite{zheng2019hfa}}&  \multirow{4}{*}{---}
                        & Dice   & 0.946 &0.893 &0.925 &0.897 &0.910 &0.964 &0.830 &0.909          \\
                    &    & IoU  &  0.898 &0.810 &0.861   &0.816 &0.836 &0.930 &0.722 &0.839     \\
                     &   & HDF   & 7.148 &33.128 &42.173 &22.903 &36.954 &12.075 &37.845 &27.461         \\
                   &     & ADB   &  0.076 &0.562 &0.210 &0.334 &0.225 &0.103 &1.685&0.456        \\\hline
            \multirow{4}{*}{$k$CBAC-Net~\cite{gu2021kcbac}} 
                      &\multirow{4}{*}{134M}  & Dice   &0.951	&0.902	&0.938	&0.911	&0.922	&0.974	&0.837 &0.919          \\
                      &  & IoU  &0.907	&0.825	&0.883	&\textbf{0.838}	&0.855	&0.949	&0.734 &0.856     \\
                     &   & HDF   &5.500	&14.940	&12.403	&15.081	&7.337	&6.848	&32.499 &13.516         \\
                      &  & ADB   &0.074	&0.285	&0.163	&0.248	&0.119	&0.059	&1.403 &0.336       \\\hline
            \multirow{4}{*}{CoTr~\cite{xie2021cotr}} 
                      &\multirow{4}{*}{42M}  & Dice   &0.944	&0.896	&0.934	&0.898	&0.908	&0.953	&\textbf{0.845} &0.911         \\
                      &  & IoU  &0.895	&0.816	&0.876	&0.816	&0.832	&0.912	&0.742 &0.841     \\
                     &   & HDF   &5.905 &15.819	&12.746	&15.767	&10.346	&12.330	&\textbf{31.539} &14.922         \\
                      &  & ADB   &0.084	&0.352	&0.159	&0.285	&0.131	&0.118	&1.391 &0.360       \\\hline
                      \multirow{4}{*}{UNETR~\cite{hatamizadeh2022unetr}} 
                      &\multirow{4}{*}{93M}  & Dice   &0.950	&0.888	&0.939	&0.903	&0.919	&0.964	&0.843 &0.915          \\
                      &  & IoU  &0.896	&0.806	&0.868	&0.818	&0.854	&0.930	&\textbf{0.744} &0.845    \\
                     &   & HDF   &5.869	&15.698	&11.769	&14.080	&8.816	&11.864	&33.392 &14.498        \\
                      &  & ADB   &0.080	&0.349	&\textbf{0.151}	&0.278	&\textbf{0.115}	&0.103	&1.404 &0.354        \\\hline
                      \multirow{4}{*}{\makecell{ConvFormer\\ (Ours)}} 
                      &\multirow{4}{*}{61M}  & Dice   &\textbf{0.955}	&\textbf{0.910}	&\textbf{0.943}	&\textbf{0.914}	&\textbf{0.926}	&\textbf{0.980}	&0.844 &\textbf{0.925}          \\
                      &  & IoU  &\textbf{0.914}	&\textbf{0.836}	&\textbf{0.885}	&0.835	&\textbf{0.857}	&\textbf{0.957}	&0.740 &\textbf{0.861}     \\
                     &   & HDF   &\textbf{5.089}	&\textbf{12.873}	&\textbf{11.745}	&\textbf{13.360}	&\textbf{7.000}	&\textbf{6.823}	&34.560 &\textbf{13.064}         \\
                      &  & ADB   &\textbf{0.069}	&\textbf{0.264}	&0.175	&\textbf{0.242}	&0.120	&\textbf{0.030}	&\textbf{1.359} &\textbf{0.323}        \\\hline
        \end{tabular}
    }
    \vspace{-4mm}
\end{table}
%


\begin{table}[t]
    \centering
    \caption{Quantitative results of different models on the 2017 ISIC skin lesion dataset.}
    \label{tab:skin}
    \scriptsize
    \resizebox{\columnwidth}{!}
    {
    \begin{tabular}{|ccccc|}
        \hline
        Method  &Jaccard Index  &Dice    &Sensitivity  &Specificity\\
        \hline
        Yuan et al.~\cite{yuan2017improving}&0.765  &0.849    &0.825  &0.975 \\ \hline
        Li et al.~\cite{li2018dense} &0.765  &0.866    &0.825  &0.984 \\ \hline
        Lei et al.~\cite{lei2020skin}&0.771  &0.859    &0.835  &0.976 \\ \hline
        Mirikharaji et al.~\cite{mirikharaji2018star} &0.773  &0.857    &0.855  &0.973 \\ \hline
        Xie et al.~\cite{xie2021sesv}&0.788  &0.868    &\textbf{0.884}  &0.957 \\ \hline
        $k$CBAC-Net~\cite{gu2021kcbac} & 0.794   &0.887  &0.847  &0.984   \\ \hline
        ConvFormer (Ours) & \textbf{0.797}$\pm$0.003   &\textbf{0.889}$\pm$0.002  &0.846$\pm$0.002  &\textbf{0.986}$\pm$0.001    \\\hline
    \end{tabular}
    }
    \vspace{-4mm}
\end{table}

\begin{table}[t]
    \centering
    \caption{Ablation study of the effects of different components in ConvFormer using the lymph node dataset.}
    \label{tab:ablation}
    \scriptsize
    \resizebox{\columnwidth}{!}
    {
    \begin{tabular}{|c|c|c|c|c|cccc|}
        \hline
       DeTrans & \makecell{Conv-based \\FFM} & \makecell{Addi.\\ E-DeTrans} &EPE  &Method  &IoU  &Precision  &Recall  &F1 \\
        \hline
        &  &  &  &  \makecell{ConvFormer \\w/o DeTrans}    &0.673   &0.833  &0.782  & 0.807   \\\hline
      $\surd$  &   &  &  &\makecell{ConvFormer \\w DeTrans}      &0.825   &0.906  &0.900  &0.903   \\\hline
       $\surd$ &  $\surd$ &  &  &\makecell{ConvFormer \\w/o \\Additional\\ Enhanced\\ DeTrans}      &0.830   &0.908  &0.906  & 0.907   \\\hline
      $\surd$& $\surd$ & $\surd$ & &\makecell{ConvFormer\\ w/o EPE}     &0.839   &0.912  &0.910  & 0.912   \\\hline
     $\surd$ & $\surd$ & $\surd$ & $\surd$ &ConvFormer     &\textbf{0.845}   &\textbf{0.925}  &\textbf{0.907}  & \textbf{0.916}   \\\hline
      \multicolumn{5}{|c|}{ConvFormer w/o residual connections}   &0.829   &0.910  &0.902  &0.906   \\\hline
    \end{tabular}
    }
    \vspace{-4mm}
\end{table}

\noindent 
{\bf Ablation Study.} 
We conduct an ablation study to examine the effects of different components in our ConvFormer using the lymph node dataset. 
As Table~\ref{tab:ablation} shows, (1) when using the fixed sinusoidal encoding~\cite{vaswani2017attention} for positional encoding (the resulted ConvFormer is denoted by ConvFormer w/o EPE), the F1 score drops by 0.4\%; 
(2) when removing the additional Enhanced DeTrans encoder that aims to explore multi-scale information from ConvFormer w/o EPE (denoted by ConvFormer w/o Additional Enhanced DeTrans), the F1 score drops by 0.5\%; 
(3) when removing Conv-based FFM from ConvFormer w/o Additional Enhanced DeTrans (denoted by ConvFormer w DeTrans), the F1 score drops by 0.4\%; 
(4) when completely removing DeTrans
(the resulted network is a pure CNN baseline, denoted by ConvFormer w/o DeTrans), the F1 score drops by 9.6\%.
These effects demonstrate the importance of our proposed components for better capturing local, global, and multi-scale information for accurate image segmentation.
Besides, we remove the residual connections of the residual-shaped hybrid stem (denoted by ConvFormer w/o residual connections), and the F1 score drops by 1.0\%. This validates the importance of the residual connections.

\vspace{-4mm}
\section{Conclusions}
\vspace{-3mm}
In this paper, we proposed a new hierarchical CNN and Transformer hybrid architecture, ConvFormer, for medical image segmentation. 
ConvFormer is capable of well learning local, global, and multi-scale representations by introducing Conv-based FFM, residual-shaped hybrid stem, and an additional Enhanced DeTrans encoder with residual connections. 
Moreover, we presented an enhanced positional encoding to improve the adaptability and flexibility of the fixed sinusoidal encoding method. 
Experiments on 2D and 3D datasets of different imaging characteristics and modalities demonstrated the effectiveness of ConvFormer.

\vspace{-2mm}
\section{Compliance with ethical standards}
\label{sec:ethics}
\vspace{-2mm}
This research study was conducted retrospectively using human subject data made available in open access by two publicly available datasets~\cite{zhuang2016multi,codella2018skin} and one in-house dataset. Ethical approval was not required as confirmed by the licenses attached with the open access datasets.



\small
\bibliographystyle{IEEEbib_abbrev}
\bibliography{refs}
\end{document}